\documentclass[lettersize,journal]{IEEEtran}
\usepackage{amsmath,amsfonts}
\usepackage{algorithm}
\usepackage{algpseudocode}
\usepackage{arydshln}
\usepackage{array}
\usepackage[caption=false,font=footnotesize,labelfont=sf,textfont=sf]{subfig}
\usepackage{hyperref}
\usepackage{cleveref}
\usepackage{textcomp}
\usepackage{stfloats}
\usepackage{url}
\usepackage{multirow}
\usepackage{verbatim}
\usepackage{graphicx}
\usepackage{caption}
\usepackage{cite}
\usepackage{booktabs}
\begin{document}

\title{Episode-specific Fine-tuning for Metric-based Few-shot Learners with Optimization-based Training}

\author{Xuanyu Zhuang, Geoffroy Peeters, Gaël Richard
\thanks{The authors are with LTCI, Télécom Paris, IP Paris, France (email: \{xuanyu.zhuang, geoffroy.peeters, gael.richard\}@telecom-paris.fr).} 
\thanks{This work was partly funded by the LISTEN-Lab, a research laboratory dedicated to machine listening from Télécom-Paris (\url{https://listen.telecom-paris.fr/en/}).}}%




\maketitle

\begin{abstract}
In few-shot classification tasks (so-called episodes), a small set of labeled support samples is provided during inference to aid the classification of unlabeled query samples. 
Metric-based models typically operate by computing similarities between query and support embeddings within a learned metric space, followed by nearest-neighbor classification. 
However, these labeled support samples are often underutilized—they are only used for similarity comparison, despite their potential to fine-tune and adapt the metric space itself to the classes in the current episode.
To address this, we propose a series of simple yet effective episode-specific, during-inference fine-tuning methods for metric-based models, including Rotational Division Fine-Tuning (RDFT) and its two variants, Iterative Division Fine-Tuning (IDFT) and Augmented Division Fine-Tuning (ADFT). 
These methods construct pseudo support-query pairs from the given support set to enable fine-tuning even for non-parametric models.
Nevertheless, the severely limited amount of data in each task poses a substantial risk of overfitting when applying such fine-tuning strategies.
To mitigate this, we further propose to train the metric-based model within an optimization-based meta-learning framework.
With the combined efforts of episode-specific fine-tuning and optimization-based meta-training, metric-based models are equipped with the ability to rapidly adapt to the limited support samples during inference while avoiding overfitting.
We validate our approach on three audio datasets from diverse domains, namely ESC-50 (environmental sounds), Speech Commands V2 (spoken keywords), and Medley-solos-DB (musical instrument). 
Experimental results demonstrate that our approach consistently improves performance for all evaluated metric-based models (especially for attention-based models) and generalizes well across different audio domains.
\end{abstract}


\section{Introduction}
By learning class-independent, transferable knowledge during training rather than class feature information, few-shot classification (FSC) models achieve generalization to unlabeled novel class samples (query set) with only few labeled examples available per class (support set) \cite{closerlook}. 
This allows FSC to overcome the limitations of scarce training data to some extent and enables effective learning for rare data. 
This capability has contributed to the growing popularity of few-shot classification (FSC) across a wide range of applications, particularly within audio-related research domains such as audio classification \cite{metaaudio, continual, classdicrimental, attgraph, multilabel}, sound event detection \cite{soundevent, soundevent_meta, active, metadcase, dcase, birdclef}, and music information retrieval \cite{musical_source, hierarchical, lcproto, drum}.

Specifically, metric-based models and optimization-based meta-learning approaches are among the most representative methods in FSC. 
Metric-based methods aim to learn an embedding space in which the similarity between query and support samples can be measured using a predefined distance function. 
Classification is then performed in a non-parametric manner, such as by applying a nearest-neighbor classifier \cite{prototypical, CAN, relation} or directly computing a weighted sum over the labels of the support samples based on similarity \cite{matchingnet}.
During inference, since the embedding space—used as the comparison metric—has already been learned, the model can directly compare query and support samples from novel classes within this space to perform classification, where the labeled support samples effectively serve as cluster points.

On the other hand, optimization-based FSC focuses on achieving efficient during-inference fine-tuning on the limited amount of labeled support set samples without overfitting. 
As a representative approach, Model-Agnostic Meta-Learning (MAML) \cite{maml} aim to learn a set of good initial model parameters that can quickly generalize to new tasks with only a few gradient updates, while its variants \cite{metasgd, meta_curvature} additionally focus on discovering better parameterized optimization strategies during fine-tuning.

In our prior work \cite{mcproto}, we investigated the feasibility of performing episode-specific fine-tuning during inference for metric-based few-shot classification models, such as the Prototypical Network (PN) \cite{prototypical}, analogous to the approach employed in optimization-based methods.
Related works have attempted similar fine-tuning strategies but achieved only marginal improvements \cite{adapte+embedding, adaptive}. 
Others have explored combining metric-based models with optimization-based approaches, yet without incorporating fine-tuning within their scope \cite{metadataset, hybrid}.
A key challenge we identified is that fine-tuning metric-based models typically requires both a labeled support set and a labeled query set to compute the loss. 
However, obtaining labeled query samples during inference is impractical, which limits the applicability of conventional fine-tuning techniques in this context.

\IEEEpubidadjcol
To this end, we proposed Rotational Division Fine-Tuning (RDFT) to manually construct pairs of labeled pseudo query sets and pseudo support sets from the given support set for fine-tuning.
Integrating RDFT, we then adopted PN as the target learner of optimization-based frameworks, in order to equip PN with the ability of fast adaption on few available support samples without overfitting.
Such approach successfully achieved a clear improvement for PN on few-shot audio classification tasks with two datasets from different audio domains, namely ESC-50 (environmental sounds) \cite{esc50} and Speech Commands V2 (spoken keywors)\cite{speech}.

Inheriting from our previous research, this paper includes the following contributions:
\begin{itemize}
    \item 
    We extend RDFT beyond Prototypical Networks (PN) by applying it to other representative metric-based models, including Matching Networks (MN) \cite{matchingnet} and Cross Attention Networks (CAN) \cite{CAN}, to demonstrate its effectiveness across diverse metric-based architectures.
    Meanwhile, a third music dataset Medley-solos-DB \cite{medley} is also introduced to further broaden the experimental evaluation.
    \item We further propose two variants of RDFT, namely Iterative Division Fine-tuning (IDFT) and Augmented Division Fine-tuning (ADFT), which modify RDFT from two opposite directions. 
    \item Specifically, for ADFT, we further explore the incorporation of audio augmentation techniques into the fine-tuning process, and investigate their particular significance for attention-based models.
\end{itemize}
The rest of the paper is organized as follows: In Section \ref{sec:preliminaries}, we discuss the working mechanisms of few-shot classification, as well as of the backbone structures that are used in this work. 
In Section \ref{sec:Proposed}, we provide the details of our proposed episode-specific fine-tuning methods and frameworks for metric-based models.
We then report the experimental settings and results, which validate our proposal, in Section \ref{sec:exp}.

\section{Preliminaries}
\label{sec:preliminaries}

\subsection{Few-shot Classification}
\label{sec:pre.few-shot}
Few-shot Classification (FSC) aims to develop classifiers that can easily generalize to previously unseen classes using only a small number of labeled samples per class. 
FSC tasks are typically formalized as $K$-way-$N$-shot episodes, where a model must classify unlabeled query samples (query set) among $K$ novel classes, each represented by $N$ labeled samples (support set).
To ensure the generalization capability of FSC models, the dataset is split by class rather than by sample. 
That is, the test set contains only samples from classes that are entirely disjoint from those used during training.

Vinyals \textit{et al.} first proposed the \textbf{episodic training principle} for few-shot learning in \cite{matchingnet}, in which the core idea is to mimic the test scenario during training. 
Specifically, training batches are constructed as $K$-way-$N$-shot episodes by randomly selecting $K$ classes and $N$ samples per class from the training set.
To maintain consistency, episodic training principle requires the same values of $K$ and $N$ being used for constructing episodes during both training and evaluation.
This training paradigm has been widely adopted and proven effective across various FSC scenarios and algorithms \cite{prototypical, attsim, CAN, meta_curvature, maml}.

\subsection{Metric-based FSC}
\label{sec:pre.metric}

\begin{table}[t]
  \caption{Symbols used throughout Section \ref{sec:preliminaries}, \ref{sec:Proposed}}
  \label{tab:symbols}
  \centering
  \renewcommand{\arraystretch}{1.1}
  \begin{tabular}{@{}ll@{}}
    \toprule
    Notations & Description \\ \midrule
    Section \ref{sec:pre.metric} & \\ \midrule
    $(S, Q)$                 & An episode consisting of labeled support set $S$ and \\
    & unlabeled query set $Q$ \\
    $K$                      & Number of classes in the support set $S$\\
    $N$                      & Number of samples per class in the support set $S$\\
    $\hat{x}$                & A query sample \\
    $\hat{y}$                & Predicted label of the query $\hat{x}$ \\
    $x_i$                    & $i^{th}$ labeled support sample in the support set $S$\\
    $y_i$                    & Ground-truth label of $x_i$ \\
    $f(\cdot)$               & Embedding network (feature extractor) \\
    $d(\cdot,\cdot)$         & Distance function (cosine / Euclidean) \\
    $c_k$                    & Prototype (mean embedding) of class $k$ \\
    $q$                      & A query embedding \\
    $w$                      & The kernel meta-leanred by the meta fusion layer \\
    $R$                      & The correlation map between $c_k$ and $q$ defined by \\
    &cosine similarity \\
    \midrule
    Section \ref{sec:pre.opt} & \\ \midrule
    $\theta$                 & Initial model parameters \\
    $\theta'$                & Copy of $\theta$ at start of inner optimization \\
    $\theta_i'$              & Adapted parameters after inner update on the $i^{th}$ episode \\
    $\alpha$                 & Inner-optimization learning rate \\
    $\beta$                  & Meta-optimization learning rate \\
    $(S_i, Q_i)$             & The $i^{th}$ sampled episode consisting of labeled support set \\
    &$S_i$ and unlabeled query set $Q_i$ \\
    $\mathcal{L}_{D}(\cdot)$ & Loss function evaluated on data $D$ \\
    $\nabla_m$               & Gradient w.r.t. $m$ \\
    $\mathbf{M}_{\mathrm{mc}}$& Learnable meta-curvature matrix \\ 
    \midrule
    Section \ref{sec:Proposed} & \\ \midrule
    $S_R,S_I,S_A,$ & The pseudo support sets and pseudo query sets constructed\\ 
    $Q_R,Q_I,Q_A$ & by RDFT, IDFT and ADFT respectively \\
    \bottomrule
  \end{tabular}
\end{table}

Metric-based FSC models learn a general embedding space that serves as a comparison metric for determining whether two samples are similar or not.
Samples from the same class are expected to be close together in the embedding space while those from different classes are well separated.\\ \\
\IEEEpubidadjcol
\textbf{Matching Network (MN)} \cite{matchingnet}. As a representative of metric-based FSC algorithms, MN predicts the query label $\hat{y}$ by calculating a weighted summation of all the support sample labels $\{y_i\}_{i=1}^{K \times N}$:
\begin{equation}
    \hat{y}=\sum_{i=1}^{K\times N} a\left(\hat{x}, x_i\right) y_i
\end{equation}
in which the weights $a\left(\hat{x}, x_i\right)$ are determined by a softmax over the cosine distance between the query sample $\hat{x}$ and each support sample $\{x_i\}_{i=1}^{K\times N}$ in the embedding space:
\begin{equation}
  a\bigl(\hat{x},x_i\bigr)
  \;=\;
  \frac{\exp\bigl(-d\bigl(f(\hat{x}),f(x_i)\bigr)\bigr)}
       {\sum_{j=1}^{K \times N}
        \exp\bigl(-d\bigl(f(\hat{x}),f(x_j)\bigr)\bigr)} .
\end{equation}
where $d$ represents the cosine distance function.
\\ \\
\textbf{Prototypical Network (PN)} \cite{prototypical}. PN further calculates the mean of support embeddings in each class as the class prototype:
\begin{equation}
    c_k=\frac{1}{N} \sum_{i=1}^N f
    \left(x_i^k\right)
\end{equation}
where $x_i^k$ stands for the $i^{th}$ support sample of class $k$.

Then a nearest neighbor classifier \cite{nn} is used to determine the query label based on the Euclidean distance between the query embedding and each prototype:
\begin{equation}
    p(y=k \mid x)=\frac{\exp \left(-d\left(f(\hat{x}), c_k\right)\right)}{\sum_{k^{\prime}=1}^K \exp \left(-d\left(f(\hat{x}), c_{k^{\prime}}\right)\right)}
\end{equation}
where $d$ stands here for the Euclidean distance function.
\\ \\
\textbf{Cross Attention Network (CAN)} \cite{CAN}. Upon the mechanism of PN, CAN introduces cross attention mechanism based on the correlation map $R\in \mathbb{R}^{m \times h \times w}$ between each pair of prototype embedding $c_k$ and query embedding $q=f(\hat{x})$ defined by cosine similarity.
The dimensions $h$ and $w$ of $R$ denote the height and width of the embedding feature maps, while $m=h \times w$.

Based on $R$, the class correlation map $R^c$ and query correlation map $R^q$ are defined by:
\begin{equation}
    R^c = R^T,
    \quad
    R^q = R
\end{equation}

After that, the class attention map $A^c\in \mathbb{R}^{h \times w}$ and query attention map $A^q\in \mathbb{R}^{h \times w}$ are generated respectively by a meta fusion layer, based on the corresponding correlation map $R^c$ and $R^q$.
Taking the class attention map $A^c$ as the example, the meta fusion layer takes $R^c$ as input and generates a meta-learned kernel $\mathbf{w} \in \mathbb{R}^{m \times 1}$:
\begin{equation}
    \mathbf{w}=W_2\left(\sigma \left(W_1\left(G A P\left(R^c\right)\right)\right.\right.
\end{equation}
in which $W_1$ and $W_2$ are the meta-learner parameters, $\sigma$ represents the RELU function \cite{relu} and $GAP$ stands for row-wise global average pooling \cite{gap}.

The generated kernel is then used to perform a convolutional fusion over each local correlation vector $\{r_i^c\}_{i=1}^m$ of $R^c$ to produce an attention scalar at the $i^{th}$ position denoted by $\{A_i^c\}_{i=1}^m$, which is further normalized by a softmax function: 
\begin{equation}
    A_i^c=\frac{\exp \left(\left(\mathbf{w}^T r_i^c\right) / \tau\right)}{\sum_{j=1}^{h \times w} \exp \left(\left(\mathbf{w}^T r_j^c\right) / \tau\right)}
\end{equation}
where $\tau$ is the temperature hyperparameter. 

The class attention map $A^c$ is then acquired by reshaping the attention scalars to a matrix in $\mathbb{R}^{h \times w}$, and further applied on the original prototype feature $c_k$:
\begin{equation}
    \bar{c}_{k} \;=\; c_{k}\;\odot\;\bigl(1 + A^{c}\bigr)
\end{equation}

Additionally, CAN is trained using a combination of two loss functions: a nearest neighbor classification loss based on distance to class prototypes, and a global classification loss which tries to classify the query embeddings among all training classes through a classifier consisting of a fully connected layer.

\subsection{Optimization-based FSC}
\label{sec:pre.opt}
Powered by meta-learning algorithms, optimization-based FSC works under the assumption that there exists an initial state of an arbitrary model that can quickly generalize to any few-shot task by fine-tuning on the support set.\\ \\
\textbf{Model-Agnostic Meta-Learning (MAML)} \cite{maml}. MAML optimizes for a set of appropriate initial model parameters of the target learner, while the target learner can be selected as an arbitrary model.
During training, MAML first randomly initializes a target learner $f_{\theta}$ and adopts a combination of an inner-optimization and a meta-optimization. 
In the inner optimization stage, the learner $f_{\theta}$ first produces a copy of its initial parameter $\theta$, namely $\theta^\prime$. Then the learner adapts $\theta^\prime$ to the labeled samples in the support set $S_i$ of the $i^{th}$ sampled episode $\{S_i, Q_i\}$, resulting in updated model parameter $\theta_i^\prime$:
\begin{equation}
    \theta_i^{\prime}=\theta^\prime-\alpha \nabla_{\theta^\prime} \mathcal{L}_{S_i}\left(f_{\theta^\prime}\right)
\end{equation}
Then during meta-optimization stage, the true initial model parameter $\theta$ is updated by the loss calculated with $\theta_i^\prime$ on the query set $Q_i$:
\begin{equation}
    \theta \leftarrow \theta-\beta \nabla_\theta \mathcal{L}_{Q_i}\left(f_{\theta_i^\prime}\right)
\end{equation}
By looping through inner-optimizations and meta-optimizations during training, the initial model parameter learns towards effective adaption on the limited amount of support set samples. Such adaption also occurs during inference, where the model quickly fine-tunes itself on the labeled support set before evaluating on the unlabeled query set. 
This also conforms to the episodic training principle where the match between training and testing scenarios is emphasized.\\ \\
\textbf{Meta-Curvature (MC)} \cite{meta_curvature}. MC is a powerful variant of MAML that does not only learn the appropriate initial parameters but also a curvature matrix $\mathbf{M_{mc}}$ which further transforms the gradient in the inner optimization stage. Therefore, the inner-optimization is now formalized as:
\begin{equation}
    \theta_i^{\prime}=\theta^\prime-\alpha \mathbf{M_{mc}} \nabla_{\theta^\prime} \mathcal{L}_{S_i}\left(f_{\theta^\prime}\right)
\end{equation}

In the meta-optimization stage, the learnable matrix  $\mathbf{M_{mc}}$ is also updated alongside the initial model parameters:
\begin{equation}
    \mathbf{M_{mc}} \leftarrow \mathbf{M_{mc}}-\beta \nabla_\mathbf{M_{mc}} \mathcal{L}_{Q_i}\left(f_{\theta_i^\prime}\right)
\end{equation}
The use of the meta-learned $\mathbf{M_{mc}}$ leads to smoother gradients during the adaptation process, making the adaptation more stable and effective compared to MAML.

\section{Proposed Method}
\label{sec:Proposed}
This section discusses two key methodologies that we propose: (1) A set of episode-specific fine-tuning methods that fine-tune metric-based FSC models on the support set of each episode during inference (sections \ref{sec:prop.rdft}, \ref{sec:prop.idft} and \ref{sec:prop.adft}) 
and (2) the corresponding training paradigms designed to equip these models with the ability to effectively adapt to the limited data available in the support set (section \ref{sec:prop.opt}).

\subsection{Motivation for Fine-tuning Metric-based FSC Models During Inference}
\label{sec:prop.motiv}
In the task setting of FSC, during inference time, a labeled support set $S$ is provided to contribute to the classification of an unlabeled query set $Q$.
Specifically, for metric-based FSC models, the labeled support samples in $S$ are embedded into a learned feature space, serving as cluster points for comparison with unlabeled query samples in a nearest neighbor classifier.

However, we consider such contribution of $S$ not enough for metric-based models, as these labeled support samples provided during inference are not directly contributing to the model parameters despite their labeled nature. 
In contrast, they could be leveraged for episode-specific fine-tuning during inference, as is done in optimization-based FSC methods.
Therefore, we propose to further leverage $S$ to fine-tune the learned feature space of metric-based FSC models during inference phase (before actually inferring the original episode $\{S, Q\}$), so that maximum use can be made of $S$. 
We emphasize that, similar to optimization-based methods, such fine-tuning is temporary and episode-specific, which means we fine-tune the base model for each inference episode independently, without retaining or accumulating the change of model parameters across different episodes.
While the embedding space of a metric-based FSC model is initially learned to distinguish between a broad set of training classes, episode-specific fine-tuning enables the model to better adapt to the certain $K$ classes sampled in a given episode \{$S, Q$\} by temporarily reshaping the embedding space.

Nevertheless, the fine-tuning phase of metric-based FSC models requires not only a labeled support set but also a labeled query set serving as the ground truth for loss calculation.
Since the collection of a labeled query set is not feasible during inference, we proposed the Rotational Division Fine-Tuning (RDFT) method in our previous work \cite{mcproto} to address this issue by manually constructing pseudo query sets and corresponding pseudo support sets from $S$.

\subsection{Rotational Division Fine-Tuning (RDFT)}
\label{sec:prop.rdft}
\begin{figure*}[!t]
\centering
\subfloat[Rotational Division Fine-Tuning (RDFT)]{\includegraphics[width=2.3in]{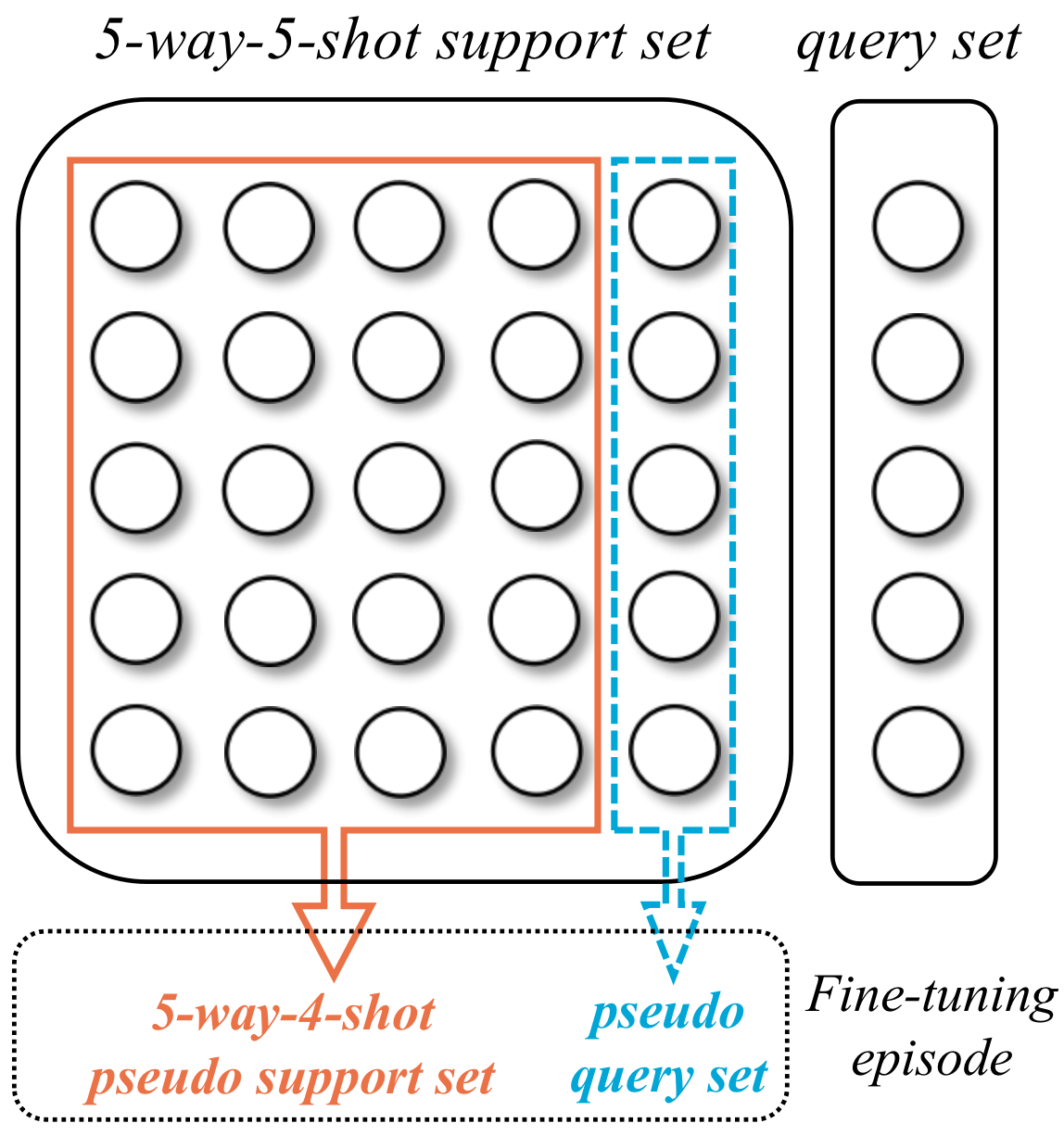}%
\label{fig_RDFT}}
\hfil
\subfloat[Augmented Division Fine-Tuning (ADFT)]{\includegraphics[width=3.2in]{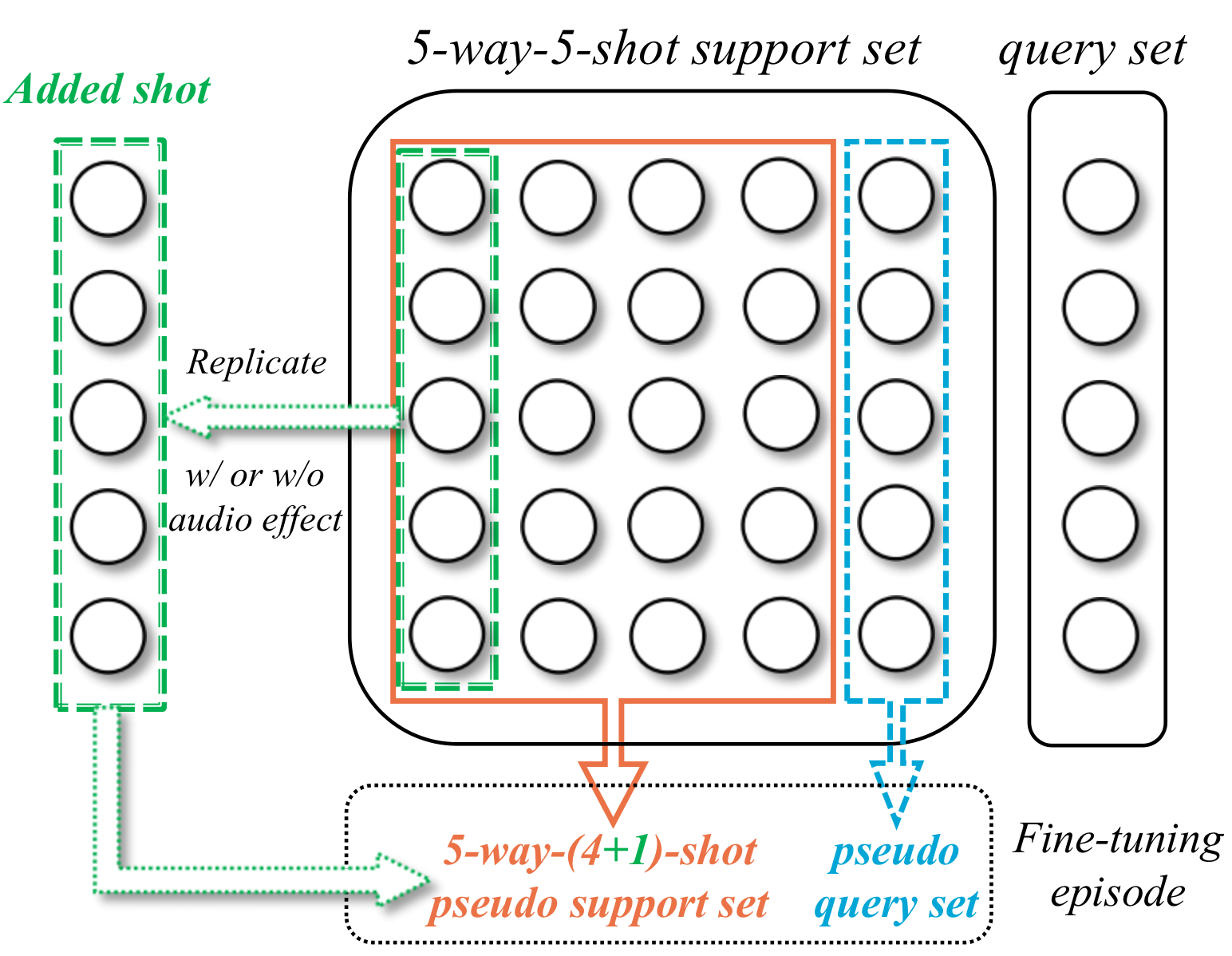}%
\label{fig_ADFT}}
\\
\subfloat[Iterative Division Fine-Tuning (IDFT)]{\includegraphics[width=7.2in]{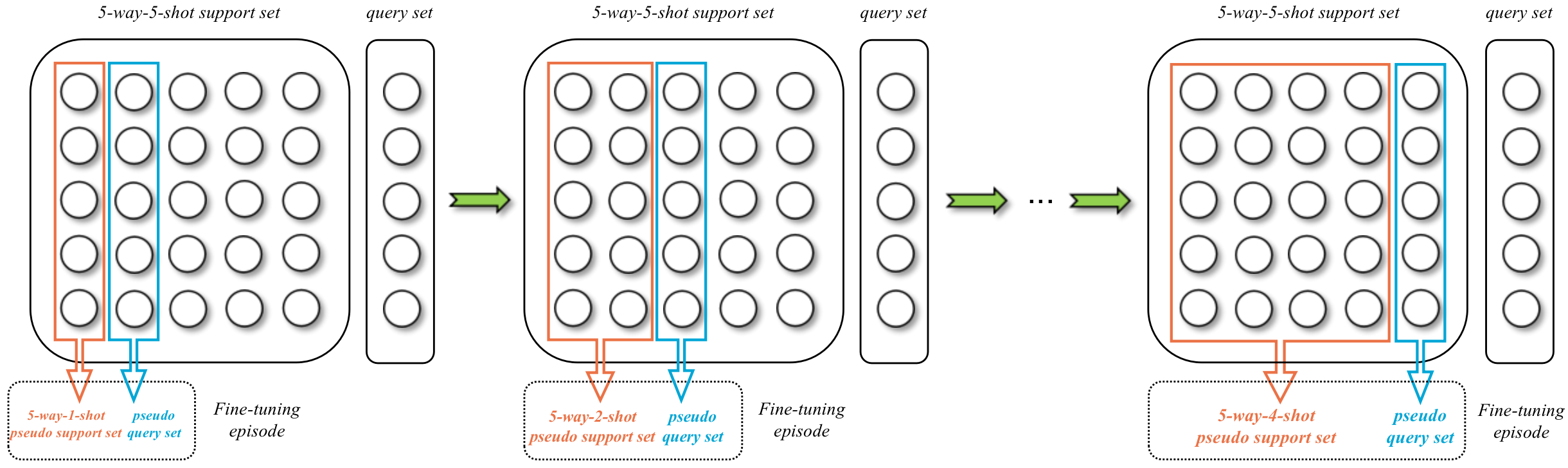}%
\label{fig_IDFT}}
\caption{The illustration of the three proposed fine-tuning methods for metric-based few-shot learners with the example of a 5-way-5-shot episode. 
\textbf{(a) Rotational Division Fine-tuning (RDFT)}. The 5-way-5-shot support set is divided into a 5-way-4-shot pseudo support set and a pseudo query set, which then together form a fine-tuning episode. Such division repeats 5 times here to make sure every column (shot) will once be selected as the pseudo query set while the rest being selected as the pseudo support set. 
\textbf{(b) Augmented Division Fine-Tuning (ADFT)}. 
Building upon RDFT, ADFT enhances each 5-way-4-shot pseudo support set generated by RDFT by adding an additional shot, resulting in a 5-way-5-shot pseudo support set. This modification maximizes adherence to the episodic training principle. The additional shot is created by sampling and replicating one shot from the existing pseudo support set, with optional audio augmentations applied to introduce more diversity into the fine-tuning process.
\textbf{(c) Iterative Division Fine-Tuning (IDFT)}. In the first iteration, the first column of the original 5-way-5-shot support set is sampled as a 5-way-1-shot pseudo support set, while the second column serves as the pseudo query set. In the second iteration, the pseudo support set expands to include the second column, and the third column is used as the new pseudo query set. This iterative expansion continues, progressively increasing the size of the pseudo support set, until the final iteration, where the original support set is divided into a 5-way 4-shot pseudo support set and a pseudo query set consisting of the last column.}
\label{fig_sim}
\end{figure*}

The core idea of RDFT is to divide the original support set $S$ to construct pairs of pseudo support and query sets, which are then used to fine-tune a metric-based FSC model during the inference stage. 
Note that this division requires the number of shots $N$ of the support set $S$ to be greater than 1; in other words, RDFT is applicable only in multi-shot scenarios.

Given a $K$-way-$N$-shot classification episode \{$S, Q$\} where $S$ and $Q$ denote the $K$-way-$N$-shot support set and the query set respectively, RDFT constructs a series of pseudo fine-tuning tasks by rotating shots within $S$.
Specifically, one shot (containing one sample per class) is extracted from $S$ to form a pseudo query set $Q_R$, while the remaining samples in $S$ constitute a $K$-way-$(N-1)$-shot pseudo support set $S_R$. 
This defines a new $K$-way-$(N-1)$-shot fine-tuning episode \{$S_R, Q_R$\}. 
This process is repeated $N$ times, each time selecting a different shot as the pseudo query set, creating $N$ distinct episodes. 
Practically, we perform multiple rounds of fine-tuning on the $N$ episodes to make sure that the model parameters are effectively updated. 
Then the updated model can be used to perform inference on the original episode \{$S, Q$\}.

In this work, we further propose two variants of RDFT, namely Iterative Division Fine-Tuning (IDFT) and Augmented Division Fine-Tuning (ADFT).
They both inherit the same underlying principle from RDFT, which is to construct a pseudo support set and a pseudo query set for fine-tuning from a given multi-shot support set $S$. 
Nevertheless, these two variants attempt to improve RDFT from different—even opposing—perspectives. 
IDFT seeks to deviate from the episodic training principle to align more closely with general machine learning rules, whereas ADFT adheres even more strictly to the episodic training principle than RDFT.
Figure \ref{fig_sim} illustrates the working mechanisms of all three fine-tuning methods in detail, using a 5-way-5-shot inference task as an example.

\subsection{Iterative Division Fine-Tuning (IDFT)}
\label{sec:prop.idft}
A limitation of RDFT method lies in its strategy for selecting the pseudo query set $Q_R$ within the original support set $S$ during each iteration. 
Specifically, except for the first iteration, the $Q_R$ selected in subsequent iterations inevitably consist of samples that have already been included in the pseudo support sets $S_R$ sampled in previous iterations. 

To solve this issue, we propose the Iterative Division Fine-Tuning (IDFT) method, whose goal is to preserve the novelty of sampled pseudo query sets across different iterations within a fine-tuning round. 
IDFT gradually expands the size of the sampled pseudo support set $S_I$ from $K$-way-$1$-shot to $K$-way-$(N-1)$-shot through iterations, as illustrated in Figure \ref{fig_IDFT}. 
By always selecting $Q_I$ with the shot (column) that is on the right to the rightmost shot of $S_I$, it is guaranteed that the model has never seen $Q_I$ in any $S_I$ sampled in previous fine-tuning iterations, thereby enhancing the validity of using such `novel' $Q_I$ as the ground truth for fine-tuning.
However, by doing this, we are actually fine-tuning the model with tasks that have various numbers of shots, which is a violation of the episodic training principle that emphasizes the match between training and testing scenario.
Therefore, IDFT essentially represents a trade-off between a general machine learning rule (to avoid the use of previously learned data as the ground truth for fine-tuning) and a few-shot learning principle (episodic training). 

\subsection{Augmented Division Fine-Tuning (ADFT)}
\label{sec:prop.adft}
Another potential flaw in RDFT is that RDFT constructs a $K$-way-$(N-1)$-shot pseudo support set $S_R$ to mimic the $K$-way-$N$-shot test scenario, but there's still a one-shot difference that prevents RDFT from fully complying with the episodic training principle. 
We further propose ADFT which simply selects one shot from the $K$-way-$(N-1)$-shot $S_R$ and replicates it, then concatenates the replicated shot back to $S_R$ to further construct a $K$-way-$N$-shot pseudo support set denoted as $S_A$. It then ensures that the fine-tuning process fully adheres to the episodic training principle (illustrated in Figure \ref{fig_ADFT}). 
Similar to RDFT, this process is repeated $N$ times, with different shots selected as the pseudo query set and the replicated shot in each iteration, thereby maintaining diversity throughout the fine-tuning process. 

We note that in each repetition, the replication operation can be regarded as simply doubling the weight of a certain shot when calculating the prototype in the embedding space.
But since every shot will once be double-weighted throughout the $N$ repetitions, the total weight of each sample remains equal.

Furthermore, we explore the use of audio augmentation methods on the replicated shot to introduce additional diversity into the sample space during fine-tuning.
We emphasize that this is not a conventional use of data augmentation, where augmentations are usually applied at scale on training data to improve a model’s generalization capability. 
We apply augmentation only to the few samples in the support set during episode-specific fine-tuning, with the goal of introducing extra diversity when reshaping the embedding space.
As a result, we do not expect our strategy to yield the same level of performance improvement typically associated with conventional data augmentation.
In fact, our strategy could instead be risky, as any instance where an augmented sample unintentionally shifts toward a different class can introduce misleading supervision, thereby negatively impacting the fine-tuning performance.
Since our experiments span over diverse audio datasets -- including music, speech, and environmental sounds -- it is essential to carefully select augmentation methods that preserve the class identity of the original samples for each audio type. 
For example, pitch shifting could be effective for environmental sounds classification but not for music instrument recognition. 
We therefore conduct a comparative analysis of the baseline `replication-only' method (the safer option) and various audio augmentation strategies across each audio dataset, as detailed in Section \ref{sec:exp.aug}.


\subsection{Optimization-based Training via Meta-learning}
\label{sec:prop.opt}
Although our proposed fine-tuning methods allow the construction of multiple different fine-tuning tasks from the original $K$-way-$N$-shot support set, the total number of available labeled samples remains fixed at $K \times N$. 
Taking a 5-way-5-shot scenario as an example in which the model is fine-tuned on only 25 labeled samples, the risk of overfitting is inherently high, while it becomes even more severe in lower-shot settings.

To address this issue, we design a training paradigm that integrates our proposed fine-tuning methods into Meta-Curvature \cite{meta_curvature}, which is an optimization-based FSC algorithm. 
As introduced in Section \ref{sec:pre.opt}, Meta-Curvature (MC) leverages meta-learning to equip a target model with the ability to fine-tune quickly and effectively on limited amount of samples in a support set while avoiding overfitting. 
Therefore, we simply adopt a metric-based FSC model as the target model of MC and replace the conventional fine-tuning approach with our proposed episode-specific fine-tuning methods.
By doing this, we also adhere to the episodic training principle fully as the metric-based FSC models perform episode-specific fine-tuning during both inference and training stage.
Specifically, the training paradigm can be described by three steps: 
\begin{enumerate}
    \item Randomly sample a $K$-way-$N$-shot episode ${(S_i, Q_i)}$ from the training data, that is to randomly select $K$ classes from the training classes and select $N$ samples from each of the chosen classes. 
    \item Before evaluating on ${(S_i, Q_i)}$, the metric-based FSC model is first fine-tuned from its initial parameters $\theta$ for $n$ steps using episode-specific fine-tuning methods on only the labeled support set $S_i$. This fine-tuning process constitutes the inner-update loop, which operates with the inner learning rate $\alpha$ and the learnable curvature matrix $\mathbf{M}$.
    \item The fine-tuned metric-based FSC model, with updated parameters $\theta_i^\prime$, is then evaluated on the original episode ${(S_i, Q_i)}$ and produces a meta-loss $\mathcal{L}_{(S_i, Q_i)}(f_{\theta_i^{\prime}})$. This meta-loss is subsequently used to update the initial parameters $\theta$ and the learnable curvature matrix $\mathbf{M}$, constituting the meta-update stage.
\end{enumerate}

Following these steps, the metric-based FSC model will optimize towards a better performance after $n$ steps of fine-tuning using our proposed episode-specific methods on the labeled support set. 
The detailed algorithmic structure is given in Algorithm \ref{alg:ADFT}, taking ADFT as the example.

\begin{algorithm}[!t]
\caption{Training Paradigm Integrating ADFT and Meta-Curvature}
\label{alg:ADFT}
\begin{algorithmic}
\State \textbf{Require:} Training episodes distribution $p(\mathcal{E})$
\State \textbf{Require:} Number of fine-tuning steps $n$
\State \textbf{Require:} Inner learning rate $\alpha$, meta learning rate $\beta$
\State \textbf{Require (optional):} Audio effect $\mathbf{T}$ for augmentation
\\
\State Initialize metric-based FSC model with parameters $\theta$
\State Initialize meta-curvature matrix $\mathbf{M}$
\While{not done}
        \State Sample ${(S_i, Q_i)} \sim p(\mathcal{E})$
        \State $\theta_i^{\prime} \leftarrow$ clone $\theta$ 
        \State $\mathbf{M}_i^{\prime} \leftarrow$ clone $\mathbf{M}$ 
        \For{step = $1$ to $n$}
            \For{j = $0$ to $size(S_i)$} 
                \State $Q_i^\prime \leftarrow \{S_{i_j}\}, \hspace{2mm} S_i^\prime \leftarrow \{\{S\} - \{Q_i^\prime\}\}$
                \State $S_{Add} \leftarrow S_{i_{j-1}}$
                \If{apply audio augmentation} 
                    \State $S_{Add} \leftarrow \mathbf{T}(S_{Add})$
                \EndIf
                \State $S_i^\prime \leftarrow Concat\{S_i^\prime, S_{Add}\}$
                \State Update $\theta_i^{\prime} \leftarrow \theta_i^{\prime}-\alpha \mathbf{M}_i^{\prime} \nabla_\theta \mathcal{L}_{(S_i^\prime, Q_i^\prime)}\left(\theta_i^{\prime}\right)$
            \EndFor{ j}
        \EndFor{ step}
        \State Update $\theta \leftarrow \theta-\beta \nabla_\theta  \mathcal{L}_{(S_i, Q_i)}\left(\theta_i^{\prime}\right)$
        \State Update $\mathbf{M} \leftarrow \mathbf{M}-\beta \nabla_\mathbf{M} \mathcal{L}_{(S_i, Q_i)}\left(\theta_i^{\prime}\right)$
\EndWhile
\end{algorithmic}
\end{algorithm}

\section{Experiments}
\label{sec:exp}
\subsection{Datasets and Pre-processing}
\label{sec:exp.dataset}
To evaluate the effectiveness of our proposed episode-specific fine-tuning methods across diverse audio domains, we conduct experiments on three widely used audio datasets: ESC-50 (environmental sounds) \cite{esc50}, Speech Commands V2 (spoken keywords) \cite{speech}, and Medley-solos-DB (music) \cite{medley}. 
For all three datasets, the audio recordings are converted into 128-bin log-scale mel-spectrograms.

\textbf{ESC-50} is a collection of 2,000 environmental audio recordings from 50 classes such as dog barking, rain, coughing, and sirens. Each class contains 40 5-second-long samples.
Following \cite{attsim}, we down-sample the clips to 16 kHz and use 35 classes for training, 5 classes for validation and 10 classes for testing.

\textbf{Speech Commands V2} \cite{speech} consists of 105,829 utterances of spoken English keywords (e.g., “yes,” “no,” “up,” “down”) spoken by different speakers.
Spanning 35 classes, the utterances are about 1-second long and are recorded by thousands of different speakers.
Version 2 improves upon the original \cite{speech1} by refining class balance and adding additional background noise classes. 
We pad or truncate the samples to fix their duration to 1 second before converting them to 128-bin log-scale mel-spectrograms.
The dataset is split into 25 classes for training, 7 for validation, and 8 for testing.

\textbf{Medley-solos-DB} \cite{medley} is a music dataset designed for instrument recognition, containing monophonic solo recordings of musical instruments across various families, such as clarinet, guitar, and violin, in which each recording is about 3 seconds in duration.
The dataset is cross-collected from the MedleyDB dataset \cite{medleyDB} and the solosDB dataset \cite{solosdb}.
Due to the limited number of instrument classes (8) in the Medley-solos-DB dataset, it is challenging to split the classes into separate training, validation, and testing sets. 
For instance, using a 4:2:2 train-val-test split and conducting 2-way classification tasks would result in the support classes of validation and test episodes always being selected as the two fixed classes, rather than randomly sampled from a broader set, which could undermine the reliability of the evaluation. 
Therefore, we simply use 4 classes for training and the remaining 4 for testing. Instead of explicitly creating a validation set, we empirically select the best model based on training performance. 
To maintain a reasonable level of task difficulty while keeping the diversity of sampled classes within each episode, we train and evaluate the model using 3-way classification tasks.

\subsection{Training Settings}
A 4-block Convolutional Neural Network (CNN) structure is used as the backbone for both PN and MN, with each of the block consisting of a convolutional layer followed by batch normalization, ReLU \cite{relu}, and a max-pooling layer.
For CAN, following the original work \cite{CAN}, a ResNet-12 network is used as the feature extractor, while adopting two convolutional layer for calculating the attention maps.

Inheriting from \cite{metaaudio}, we set inner-optimization learning rate $\alpha$ to 0.2 for ESC-50 and Medley-solos-DB, and to 0.02 for Speech Commands V2 dataset. 
The meta-optimization learning rate $\beta$ is set to $1 \times 10^{-3}$ for all three datasets.
The models are fine-tuned by our proposed episode-specific fine-tuning methods for 8 gradient steps on each sampled episode during inner-optimization of both training and inference phases.

All experiments are implemented using Pytorch \cite{pytorch}. For implementations of meta-learning algorithms like MC, we used the learn2learn Python package \cite{l2l}.
The code examples are open-sourced at \url{https://github.com/zdsy/Episode-specific-FT}.

\subsection{Effect of RDFT on Different Metric-based Models}
\label{sec:exp.rdft}
We have shown in our previous work \cite{mcproto} that 
RDFT effectively improves the performance of Prototypical Networks (PN) in few-shot audio classification tasks on ESC-50 and Speech Commands V2.
We also proposed that RDFT theoretically should benefit all metric-based FSC models besides PN.
Therefore, we further investigate the impact of RDFT on other metric-based models including Matching Network (MN) and Cross Attention Network (CAN) on ESC-50, Speech Commands V2 and the newly introduced music dataset Medley-solos-DB.
Table \ref{tab:esc50}, Table \ref{tab:scv2} and Table \ref{tab:medley} indicates the results on these three datasets respectively, in the form of average classification accuracy (ACC) with a 95\% confidence interval.
Specifically, we experiment 3-way-5-shot tasks on Medley-solos-DB due to the class volume limitation, and conduct 5-way-5-shot tasks on the rest of the two datasets.
The vanilla PN, MN and CAN are used as baseline for comparisons with our proposed RDFT frameworks integrating Meta-Curvature (MC).
We report the results both before and after fine-tuning, and quantify the performance gain as the difference in accuracy between the baseline model and its fine-tuned RDFT variant.

\begin{table}[t]
    \caption{Average classification accuracy (with 95\% confidence interval) of \textbf{5-way-5-shot environmental sound event classification on ESC-50}.}
    \centering
    \begin{tabular}{c|c c|c}
    \toprule
        \textbf{Model}&\textbf{w/o fine-tuning}&\textbf{w/ fine-tuning}&\textbf{Gain}  \\
        \midrule
        PN&$82.34\pm0.18$\%&--&--\\
        MC-PN-RDFT&$84.12\pm0.17$\% &\textbf{85.97 $\pm$ 0.16\%} &+3.63\%  \\
        \midrule
        MN&$81.54\pm0.15$\%&--&-- \\
        MC-MN-RDFT&$82.64\pm0.15$\% &\textbf{83.63 $\pm$ 0.14\%} &+2.09\%  \\
        \midrule
        CAN&83.01 $\pm$ 0.14\%&--&-- \\
        MC-CAN-RDFT&81.67 $\pm$ 0.15\%&\textbf{88.23 $\pm$ 0.12\%}&+\textbf{5.22\%} \\
    \bottomrule
    \end{tabular}
    \label{tab:esc50}
\end{table}

\begin{table}[t]
    \caption{Average classification accuracy (with 95\% confidence interval) of \textbf{5-way-5-shot keyword detection on Speech Commands V2}.}
    \centering
    \begin{tabular}{c|c c|c}
    \toprule
        \textbf{Model}&\textbf{w/o fine-tuning}&\textbf{w/ fine-tuning}&\textbf{Gain}  \\
        \midrule
        PN&$86.10\pm0.14$\%&--&--\\
        MC-PN-RDFT&$86.84\pm0.13$\%&\textbf{86.94 $\pm$ 0.13\%}&+0.84\%  \\
        \midrule
        MN&$77.91\pm0.16$\%&--&-- \\
        MC-MN-RDFT&$79.49\pm0.16$\%&\textbf{81.29 $\pm$ 0.15\%}&+3.38\%  \\
        \midrule
        CAN&74.92 $\pm$ 0.17\%&--&-- \\
        MC-CAN-RDFT&75.12 $\pm$ 0.17\%&\textbf{82.88 $\pm$ 0.15\%}&+\textbf{7.96\%} \\
    \bottomrule
    \end{tabular}
    \label{tab:scv2}
\end{table}

\begin{table}[!t]
    \caption{Average classification accuracy (with 95\% confidence interval) of \textbf{3-way-5-shot instrument recognition on Medley-solos-DB.}}
    \centering
    \begin{tabular}{c|c c|c}
    \toprule
        \textbf{Model}&\textbf{w/o fine-tuning}&\textbf{w/ fine-tuning}&\textbf{Gain}  \\
        \midrule
        PN&$80.27\pm0.25$\%&--&--\\
        MC-PN-RDFT&$81.20\pm0.25$\%&\textbf{81.95 $\pm$ 0.23\%}&+1.68\%  \\
        \midrule
        MN&$84.90\pm0.18$\%&--&-- \\
        MC-MN-RDFT&$84.01\pm0.15$\%&\textbf{85.85 $\pm$ 0.17\%}&+0.95\%  \\
        \midrule
        CAN&80.02 $\pm$ 0.20\%&--&-- \\
        MC-CAN-RDFT&73.14 $\pm$ 0.22\%&\textbf{84.85 $\pm$ 0.18\%}&+\textbf{4.83\%} \\
    \bottomrule
    \end{tabular}
    \label{tab:medley}
\end{table}

The results successfully demonstrate that our proposed method introduces benefit for all three tested metric-based models on all three datasets.
We also notice that the effectiveness of RDFT is related to the number of samples in the support set of each episode, as a larger support set generally carries richer class information, leading to more reliable class feature estimation and less risk of overfitting.
As a result, the performance gain brought by RDFT on Medley-solos-DB (15 support samples per episode) is averagely smaller than on the other two datasets (25 support samples per episode). 

\begin{figure}[t]
    \centering
    \includegraphics[width=9cm]{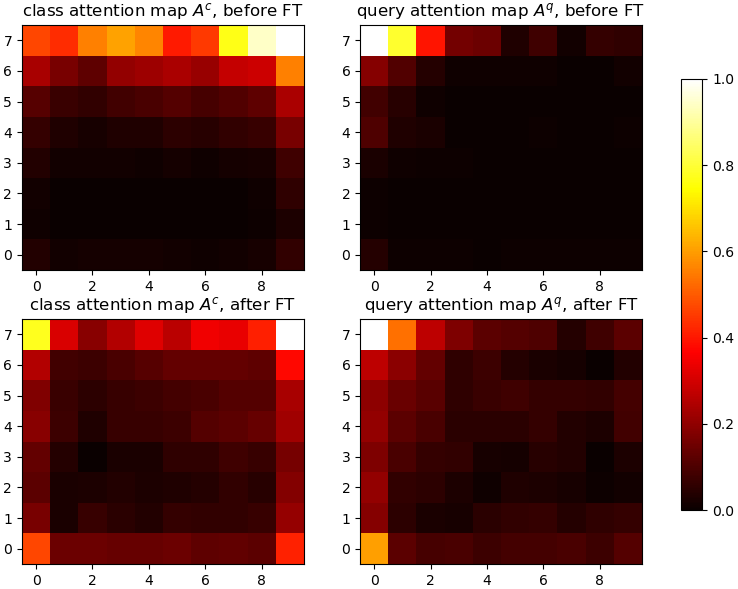}
    \caption{The cross attention map of CAN on the extracted feature space before and after episode-specific fine-tuning, where x-axis and y-axis represents the two principal components of the features. The top row represents the attention map of the initial model state before fine-tuning, giving an average classification accuracy of \textbf{80.89 $\pm$ 0.15\%} on 5-way-5-shot environmental sound classification tasks with ESC-50. The bottom row represents the attention map of the fine-tuned model, giving an increased accuracy of \textbf{88.01 $\pm$ 0.12\%}.}
    \label{fig:att_map}
\end{figure}

Moreover, we observe that RDFT yields significantly larger performance improvement for CAN compared to the other two non-attention-based models. 
Additionally, for MC-CAN-RDFT, the performance gap before and after fine-tuning is noticeably larger. 
We hypothesize that this is because RDFT enables attention-based models like CAN to temporarily adapt not only their embedding space but also their attention maps, allowing the model to better focus on the classes present in the current episode.
To validate this hypothesis, we visualize the attention maps of MC-CAN-RDFT before and after fine-tuning in Figure \ref{fig:att_map}. 
The results show that the cross attention maps of the initial model before fine-tuning, illustrated by the two sub-figures in the top row, are primarily concentrated on top regions of the feature space, leaving much of the feature content underutilized. 
In contrast, after fine-tuning, the attention weights are more evenly distributed across the entire feature space, leading to a more reasonable and comprehensive focus.
This provides an intuitive explanation for the large performance gain observed after fine-tuning.

\subsection{Comparative Analysis of RDFT, IDFT and ADFT}
\label{sec:exp.compare}
\begin{table*}[!t]
    \caption{Average classification accuracy (with 95\% confidence interval) on 5-way-5-shot (ESC-50, Speech Commands V2) and 3-way-5-shot (Medley-solos-DB) audio classification tasks.}
    \centering
    \begin{tabular}{c|cc|cc|cc}
        \toprule
        \multirow{2}{*}{\textbf{Models}}
        & \multicolumn{2}{c|}{\textbf{ESC-50}} 
        & \multicolumn{2}{c|}{\textbf{Speech Commands V2}} 
        & \multicolumn{2}{c}{\textbf{Medley-solos-DB}} \\
        \cmidrule(r){2-3} \cmidrule(r){4-5} \cmidrule(r){6-7}
        & \textbf{w/o FT} & \textbf{w/ FT} 
        & \textbf{w/o FT} & \textbf{w/ FT} 
        & \textbf{w/o FT} & \textbf{w/ FT} \\
        \midrule
        PN & $82.34\pm0.18$\% & -- & $86.10\pm0.14$\% & -- & $80.27\pm0.25$\% & -- \\
        MC-PN-IDFT & $83.11\pm0.18$\% & $84.43\pm0.17$\% & $86.55\pm0.13$\% & $86.63\pm0.13$\% & $80.04\pm0.24$\% & $81.95\pm0.23$\% \\
        MC-PN-RDFT & $84.12\pm0.17$\% & $85.97\pm0.16$\% & $86.84\pm0.13$\% & $86.94\pm0.13$\% & $81.20\pm0.25$\% & $83.93\pm0.22$\% \\
        MC-PN-ADFT & 85.68 $\pm$ 0.17\% & \textbf{87.08 $\pm$ 0.16\%} & 87.02 $\pm$ 0.13\% & \textbf{87.07 $\pm$ 0.13\%} & 81.93 $\pm$ 0.27\% & \textbf{84.32 $\pm$ 0.24\%} \\
        \midrule
        MN & $81.54\pm0.15$\% & -- & $77.91\pm0.16$\% & -- & $84.90\pm0.18$\% & -- \\
        MC-MN-IDFT & $81.47\pm0.15$\% & $82.83\pm0.14$\% & $78.64\pm0.16$\% & $80.27\pm0.16$\% & $83.81\pm0.18$\% & $85.17\pm0.18$\% \\
        MC-MN-RDFT & $82.64\pm0.15$\% & $83.63\pm0.14$\% & $79.49\pm0.16$\% & $81.29\pm0.15$\% & $84.01\pm0.18$\% & $85.85\pm0.17$\% \\
        MC-MN-ADFT & 83.36 $\pm$ 0.14\% & \textbf{84.53 $\pm$ 0.14\%} & 81.36 $\pm$ 0.15\% & \textbf{83.00 $\pm$ 0.15\%} & 84.53 $\pm$ 0.18\% & \textbf{86.10 $\pm$ 0.17\%} \\
        \midrule
        CAN & $83.01\pm0.14$\% & -- & $74.92\pm0.17$\% & -- & $80.02\pm0.20$\% & -- \\
        MC-CAN-IDFT & $80.57\pm0.15$\% & $87.46\pm0.13$\% & $76.86\pm0.16$\% & $82.25\pm0.15$\% & $68.45\pm0.23$\% & $81.22\pm0.20$\% \\
        MC-CAN-RDFT & $81.67\pm0.15$\% & \textbf{88.23 $\pm$ 0.12\%} & $75.12\pm0.17$\% & \textbf{82.88 $\pm$ 0.15\%} & $75.81\pm0.20$\% & $84.84\pm0.18$\% \\
        MC-CAN-ADFT & 80.89 $\pm$ 0.15\% & 88.01 $\pm$ 0.12\% & 76.76 $\pm$ 0.16\% & 82.74 $\pm$ 0.15\% & 73.14 $\pm$ 0.22\% & \textbf{84.85 $\pm$ 0.18\%} \\
        \bottomrule
    \end{tabular}
\label{tab:merged_finetuning}
\end{table*}

We further conduct a set of experiments to validate the two variants of our proposed RDFT method, namely IDFT and ADFT. 
All model architectures and training settings remain consistent with the RDFT experiments, with the only difference being the episode-specific fine-tuning strategy employed.
The experimental results are presented in Table \ref{tab:merged_finetuning}. 

As discussed in Section \ref{sec:Proposed}, IDFT deviates from the episodic training principle and instead aligns more closely with the general machine learning paradigm, while ADFT adheres more strictly to the episodic training principle. 
Although ADFT relies on simple sample replication, which may introduce sample-wise biases during fine-tuning, it generally outperforms RDFT in most cases—except when applied to CAN-based models. 
In contrast, IDFT consistently underperforms relative to RDFT across all tasks.
These results strongly highlight the importance of preserving episodic consistency when training FSC models.
As for CAN, since the attention map is driven by the cosine similarity matrix between the query embedding and each class prototype, it is required that the prototypes are representative enough by carrying varied local patterns.
Due to this, simply replicating one sample with ADFT shifts the prototype toward that sample and could lower the spectral variance of the matrix, leading to a locally-biased attention map.
In Section \ref{sec:prop.adft}, we stated that although each iteration of ADFT may locally shift the embedding space toward the replicated sample, the overall expectation of the embedding space remains unbiased with respect to any specific sample by repeating ADFT.
However, for the learnable attention map which takes the cosine similarity matrix as input, it is more sensitive to the local biases introduced by duplicating specific samples, making it more difficult to eliminate the negative effects.
Together, these help explain why ADFT consistently outperforms RDFT on non-attention-based models (PN, MN) in various tasks while underperforming on CAN, according to Table \ref{tab:merged_finetuning}.
Further experimental results (details given in Section \ref{sec:exp.aug-can}) also support the above hypothesis, that adding audio augmentations on the replicated sample restores the variance and thus made ADFT outperforms RDFT on CAN. 

\subsection{Audio Effects in ADFT for non-attention-based model (PN)}
\label{sec:exp.aug}
We further experimented with applying waveform-level audio augmentation to the replicated samples in ADFT to avoid exact duplication and restore intra-class variance.
As mentioned in Section \ref{sec:prop.adft}, we experimented a range of audio augmentation methods on all three datasets to determine the most suitable method for each dataset. 
Specifically, we use the MC-PN-ADFT model in these experiments, as PN is a representative non-attention-based model and is straightforward to implement.
Controlled by a \texttt{scale} hyperparameter $s$, the candidate audio effects we evaluate include the following:
\begin{itemize}
    \item \textbf{Replication-only (no augmentation)} as the baseline.
    \item \textbf{Additive noise ($s=1$).} Generated random-colored noise is added to the replicated audio sample after being rescaled to a uniformly sampled signal-to-noise ratio (SNR) in $[12,100]\,\mathrm{dB}$. The noise is shaped in the frequency domain with a randomly sampled power spectral density profile.
    \item \textbf{Frequency filter ($s=4$).} Multiple band-pass filters with center frequencies randomly sampled from $[30, 3000]\,\mathrm{Hz}$ and gains from $[-8, 8]\,\mathrm{dB}$ are applied using \texttt{torchaudio.functional.equalizer\_biquad} function \cite{pytorch}, altering the spectral characteristics of the audio.
    \item \textbf{Pitch shifting ($s=4$).} The pitch of the replicated audio sample is shifted up or down by a random number of semitones within a range of $\{-2, -1, 1, 2\}$, using \texttt{librosa.effects.pitch\_shift} function \cite{librosa}. 
    \item \textbf{Random augmentation.} In each repetition of ADFT, one of the three augmentation methods listed above is randomly selected and applied to the replicated sample.
\end{itemize}

According to the experimental results shown in Table \ref{tab:aug}, we observe that pitch shifting works best for ESC-50, while a frequency filter produces best results for both Speech Commands V2 and Medley-solos-DB.
\begin{table*}[!t]
    \caption{Comparative analysis of applying various audio augmentation methods in ADFT across different audio datasets, measured by the average classification accuracy (with 95\% confidence intervals) on 5-way-5-shot (ESC-50, Speech Commands V2) and 3-way-5-shot (Medley-solos-DB) audio classification tasks using \textbf{MC-PN-ADFT}.}
    \centering
    \begin{tabular}{c|cc|cc|cc}
        \toprule
        \multirow{2}{*}{\textbf{Audio effects}}
        & \multicolumn{2}{c|}{\textbf{ESC-50}} 
        & \multicolumn{2}{c|}{\textbf{Speech Commands V2}} 
        & \multicolumn{2}{c}{\textbf{Medley-solos-DB}} \\
        \cmidrule(r){2-3} \cmidrule(r){4-5} \cmidrule(r){6-7}
        & \textbf{w/o FT} & \textbf{w/ FT} 
        & \textbf{w/o FT} & \textbf{w/ FT} 
        & \textbf{w/o FT} & \textbf{w/ FT} \\
        \midrule
        Replication-only (no augmentation) & $85.68\pm0.17$\% & $87.08\pm0.16$\% & $87.02\pm0.13$\% & $87.07\pm0.13$\% & $81.93\pm0.27$\% & $84.32\pm0.24$\% \\ 
        \noalign{\vskip 2pt}
        \cdashline{1-7}
        \noalign{\vskip 3pt}
        Additive noise & $83.82\pm0.18$\% & $86.35\pm0.17$\% & $87.13\pm0.13$\% & $87.18\pm0.13$\% & $80.46\pm0.27$\% & $83.13\pm0.24$\% \\
        Frequency filter & $84.80\pm0.17$\% & $86.71\pm0.16$\% & $87.31\pm0.13$\% & \textbf{87.34 $\pm$ 0.13\%} & $81.50\pm0.23$\% & \textbf{84.36 $\pm$ 0.21\%} \\
        Pitch shifting & $86.06\pm0.16$\% & \textbf{87.12 $\pm$ 0.16}\% & $87.16\pm0.13$\% & $87.17\pm0.13$\% & $81.07\pm0.26$\% & $83.59\pm0.23$\% \\
        Random augmentation & $84.59\pm0.18$\% & $86.54\pm0.16$\% & $86.95\pm0.13$\% & $86.98\pm0.13$\% & $81.35\pm0.27$\% & $84.07\pm0.23$\% \\
        \bottomrule
    \end{tabular}
\label{tab:aug}
\end{table*}

\begin{figure}[!t]
    \centering
    \includegraphics[width=8.8cm]{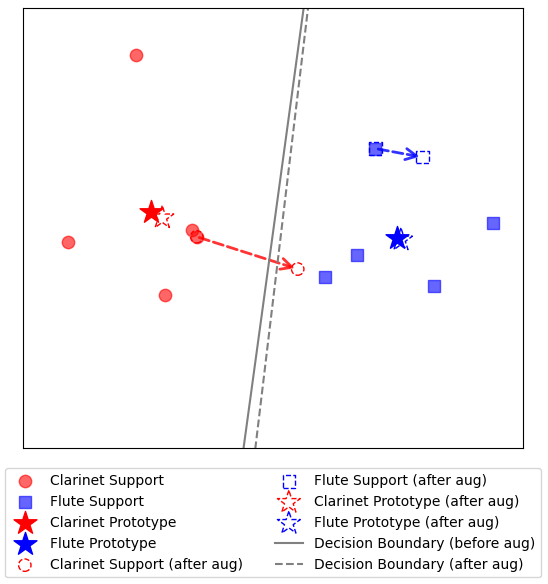}
    \caption{An illustration of how audio augmentation changes the location of the support samples and prototypes in the learned embedding space of a vanilla PN \cite{prototypical}, with arrows indicating the moving direction.}
    \label{fig:shift}
\end{figure}

Nevertheless, even if an appropriate augmentation is chosen, applying audio augmentation to the replicated samples of ADFT yields only marginal improvements over the baseline model without augmentation. 
Some augmentations are even harmful when applying to specific datasets.
A notable example is the use of pitch shifting on the Medley-solos-DB dataset, which resulted in approximately a 1\% drop in accuracy compared to the replication-only MC-PN-ADFT model. 
This outcome is reasonable, as pitch shifting in instrument recognition tasks can alter the original timbral characteristics of the instrument, thereby confusing the classification model.
For example, by shifting the pitch of clarinet samples upward or flute samples downward, their features in the embedding space may become less distinguishable. 
To illustrate this, we conducted a 2-way-5-shot classification experiment on these two classes using pitch-up augmentation to analyze the position changes of the support samples and class prototypes in the embedding space before and after augmentation.
We illustrate this in Figure \ref{fig:shift}, where the red and blue dots represent support samples from clarinet and flute respectively, while red and blue stars stands for their corresponding class prototypes.
Dashed symbols represent the augmented support samples and prototypes, with arrows indicating their moving directions. 
The decision boundaries before and after augmentation are also exhibited. 
The visualization clearly shows that pitch-shifted clarinet samples move significantly closer to the flute prototype in the embedding space, even crossing the decision boundary, thereby increasing the risk of misclassification.
Despite this example being somewhat extreme, similar label-confusing scenarios will likely occur across thousands of randomly sampled training episodes, and can accumulate to degrade overall model performance.

In summary, for non-attention-based models such as PN, introducing well-chosen 
audio augmentations in ADFT can provide slight performance gains by avoiding exact duplication and restoring intra-class variance. Though, for other less appropriate augmentations, 
the baseline replication-only strategy could be a relatively safer and more stable option.

\begin{table*}[t!]
    \caption{Average classification accuracy (with 95\% confidence interval) on 5-way-5-shot (ESC-50, Speech Commands V2) and 3-way-5-shot (Medley-solos-DB) audio classification tasks using \textbf{CAN}-based models. For MC-CAN-ADFT (Aug), the best augmentation method is chosen for each dataset according to the results in Table \ref{tab:aug}.}
    \centering
    \begin{tabular}{c|cc|cc|cc}
        \toprule
        \multirow{2}{*}{\textbf{Models}}
        & \multicolumn{2}{c|}{\textbf{ESC-50}} 
        & \multicolumn{2}{c|}{\textbf{Speech Commands V2}} 
        & \multicolumn{2}{c}{\textbf{Medley-solos-DB}} \\
        \cmidrule(r){2-3} \cmidrule(r){4-5} \cmidrule(r){6-7}
        & \textbf{w/o FT} & \textbf{w/ FT} 
        & \textbf{w/o FT} & \textbf{w/ FT} 
        & \textbf{w/o FT} & \textbf{w/ FT} \\
        \midrule
        CAN & $83.01\pm0.14$\% & -- & $74.92\pm0.17$\% & -- & $80.02\pm0.20$\% & -- \\
        MC-CAN-IDFT & $80.57\pm0.15$\% & $87.46\pm0.13$\% & $76.86\pm0.16$\% & $82.25\pm0.15$\% & $68.45\pm0.23$\% & $81.22\pm0.20$\% \\
        MC-CAN-RDFT & $81.67\pm0.15$\% & 88.23 $\pm$ 0.12\% & $75.12\pm0.17$\% & 82.88 $\pm$ 0.15\% & $75.81\pm0.20$\% & $84.84\pm0.18$\% \\
        MC-CAN-ADFT & 80.89 $\pm$ 0.15\% & 88.01 $\pm$ 0.12\% & 76.76 $\pm$ 0.16\% & 82.74 $\pm$ 0.15\% & 73.14 $\pm$ 0.22\% & 84.85 $\pm$ 0.18\% \\
        \textbf{MC-CAN-ADFT (Aug)} & 81.30 $\pm$ 0.15\% & \textbf{88.70 $\pm$ 0.12\%} & 76.37 $\pm$ 0.17\% & \textbf{83.50 $\pm$ 0.14\%} & 73.74 $\pm$ 0.22\% & \textbf{85.11 $\pm$ 0.18\%} \\
        \bottomrule
    \end{tabular}
\label{tab:CAN_aug}
\end{table*}

\subsection{Audio Effects in ADFT for attention-based model (CAN)}
\label{sec:exp.aug-can}

The impact of applying audio augmentations in ADFT is however different for attention-based models such as CAN.
As discussed in Section \ref{sec:exp.compare}, the attention map in CAN is particularly sensitive to the diversity of local patterns within class features.
As a result, the replication-only version of ADFT underperforms even RDFT on CAN, highlighting the limitations of simple duplication in this context. 
In such cases, the ability of audio augmentation to restore feature variance becomes especially beneficial.

Based on Table \ref{tab:aug}, we selected the best-performing audio effect for each dataset (pitch shifting for ESC-50 and frequency filter for Speech Commands V2 \& Medley-solos-DB) and applied it to MC-CAN-ADFT. 
With the results shown in Table \ref{tab:CAN_aug}, it is demonstrated that incorporating audio effects on the replicated samples in MC-CAN-ADFT leads to a considerable performance boost, surpassing MC-CAN-RDFT. 
This further validates our earlier hypothesis, that for attention-based models like CAN, although replication-only ADFT strictly adheres to the episodic fine-tuning principle, the loss of variance can still degrade model performance. 
However, by applying appropriate audio effects to the replicated samples, the variance can be restored and allows the model to fully benefit from the adherence of the episodic training principle.

\section{Conclusion}
This work presented a set of novel episode-specific fine-tuning strategies to enhance the performance of metric-based few-shot classification models during inference, paired with an optimization-based meta-training framework to improve the efficiency of fine-tuning. 
Extensive evaluations across three distinct audio datasets and metric-based architectures demonstrated the effectiveness and generalizability of the proposed methods. 
Notably, the methods yielded significant improvements for attention-based models by jointly adapting both the embedding space and the attention mechanisms to the specific classes of each episode. 
Among all proposed strategies, Augmented Division Fine-Tuning (ADFT) proved to be the most effective, consistently yielding the greatest performance gains for non-attention-based models, as well as for attention-based models when appropriately selected audio augmentations are incorporated.

Future research will focus on establishing principled methodologies for selecting optimal augmentation strategies to further enhance model robustness and domain adaptability.

\bibliographystyle{IEEEtran}
\bibliography{refs}

\vfill

\end{document}